\documentclass[sigconf]{acmart}
\AtBeginDocument{%
  \providecommand\BibTeX{{%
    \normalfont B\kern-0.5em{\scshape i\kern-0.25em b}\kern-0.8em\TeX}}}

\usepackage[utf8]{inputenc}
\usepackage{microtype}
\usepackage{multirow}
\usepackage{subfigure}
\usepackage[utf8]{inputenc}
\usepackage{graphicx}
\usepackage{amsmath}
\usepackage{algorithm}
\usepackage{algorithmic}
\usepackage{CJKutf8}
\usepackage{balance}
\usepackage[skip=2.5pt]{caption}
\usepackage{array}
\usepackage{booktabs}
\usepackage{threeparttable}
\usepackage{soul}
\usepackage{setspace}

\copyrightyear{2024}
\acmYear{2024}
\setcopyright{acmlicensed}
\acmDOI{10.1145/3589335.3648302}
\acmISBN{979-8-4007-0172-6/24/05}

\begin{document}

\title{
A Semi-supervised Multi-channel Graph Convolutional Network for Query Classification in E-commerce 
}




\author{Chunyuan Yuan}
\email{yuanchunyuan1@jd.com}
\affiliation{
  \institution{JD.COM}
  \city{Beijing}
  \country{China}
}

\author{Ming Pang}
\email{pangming8@jd.com}
\affiliation{
  \institution{JD.COM}
  \city{Beijing}
  \country{China}
}

\author{Zheng Fang}
\email{fangzheng21@jd.com}
\affiliation{
  \institution{JD.COM}
  \city{Beijing}
  \country{China}
}

\author{Xue Jiang}
\email{jiangxue@jd.com}
\affiliation{
  \institution{JD.COM}
  \city{Beijing}
  \country{China}
}

\author{Changping Peng}
\email{pengchangping@jd.com}
\affiliation{
  \institution{JD.COM}
  \city{Beijing}
  \country{China}
}

\author{Zhangang Lin}
\email{linzhangang@jd.com}
\affiliation{
  \institution{JD.COM}
  \city{Beijing}
  \country{China}
}

\renewcommand{\shortauthors}{Chunyuan Yuan, et al.}

\begin{abstract}
Query intent classification is an essential module for customers to find desired products on the e-commerce application quickly. Most existing query intent classification methods rely on the users' click behavior as a supervised signal to construct training samples. However, these methods based entirely on posterior labels may lead to serious category imbalance problems because of the Matthew effect in click samples. Compared with popular categories, it is difficult for products under long-tail categories to obtain traffic and user clicks, which makes the models unable to detect users' intent for products under long-tail categories. This in turn aggravates the problem that long-tail categories cannot obtain traffic, forming a vicious circle. In addition, due to the randomness of the user's click, the posterior label is unstable for the query with similar semantics, which makes the model very sensitive to the input, leading to an unstable and incomplete recall of categories.

In this paper, we propose a novel \textbf{S}emi-supervised \textbf{M}ulti-channel \textbf{G}raph \textbf{C}onvolutional \textbf{N}etwork (SMGCN) to address the above problems from the perspective of label association and semi-supervised learning. SMGCN extends category information and enhances the posterior label by utilizing the similarity score between the query and categories. Furthermore, it leverages the co-occurrence and semantic similarity graph of categories to strengthen the relations among labels and weaken the influence of posterior label instability. We conduct extensive offline and online A/B experiments, and the experimental results show that SMGCN significantly outperforms the strong baselines, which shows its effectiveness and practicality. 
\end{abstract}

\ccsdesc[500]{Information systems~Query intent}
\ccsdesc[500]{Computing methodologies~Natural language processing}

\keywords{
Multi-label Text Classification,
Query Intent Classification, 
Semi-supervised Learning, 
E-commerce Retrieval
}




\maketitle

\section{Introduction}
Online shopping has evolved into a fundamental aspect of our lives, significantly reshaping our daily routines over the past few years. An increasing number of e-commerce platforms such as Amazon, Taobao, and JD offer customers hundreds of millions of vibrant and colorful products. These massive products are organized in the form of categories to facilitate customers to retrieve them quickly. To cover as many kinds of commodities as possible, the category taxonomy involves nearly ten thousand leaf categories in e-commerce applications. Due to the diversity of user needs and plenty of categories, accurately capturing the user's intention to purchase the category of products is a crucial part of the e-commerce platform.

\begin{figure}[htbp]
	\centering
	\subfigure[Query rank versus count.]{
		\label{fig:zipf} 
		\includegraphics[width=0.22\textwidth]{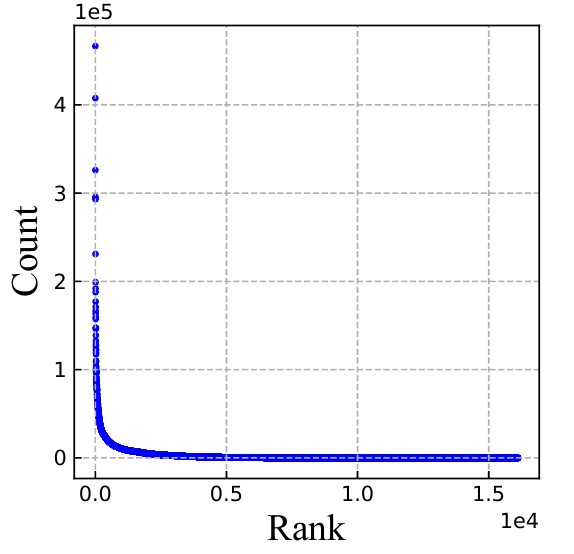}
	}
	\subfigure[Logarithmic curve of (a).]{
		\label{fig:zipf_log} 
		\includegraphics[width=0.21\textwidth]{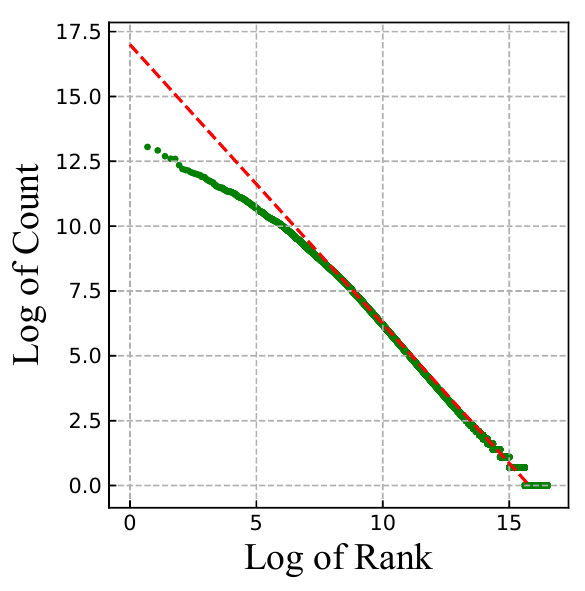}
        }
	\caption{Zipf’s law: query rank versus search count distribution.}
	\label{fig:long-tail}
\end{figure}

Query intent classification has gained significant attention from both academia and industry. Early research uses click graphs~\cite{li2008learning} or context information~\cite{cao2009context} to solve the short and ambiguous problems of query faced by the general Web search. In recent years, query intent classification has usually been regarded as a multi-label text classification problem in academia. With the wide application of deep learning technology, some deep learning-based models, such as XML-CNN~\cite{liu2017deep}, KRF~\cite{ma2020beyond}, HiAGM~\cite{zhou2020hierarchy}, LSAN~\cite{xiao2019label} have been proposed to learn the contextual information of documents to enhance the representation learning of queries. Furthermore, some recently proposed query intent classification models, such as PHC~\cite{zhang2019improving}, DPHA~\cite{zhao2019dynamic}, and MMAN~\cite{yuan2023multi} also explore utilizing the correlation between query intent classification and semantic textual similarity or multi-task to facilitate models to learn external information beyond query information.

Most previous methods assume an abundance of authentic labeled data is available to train a model. However, manual annotation is expensive and time-consuming, especially for the thousands of product categories. As a result, the industry often utilizes users' click behavior as an implicit feedback signal to generate training samples, but this approach has its challenges. One major issue is the category imbalance in the training data, where long-tail categories struggle to obtain user clicks and traffic, making it difficult for models to identify them. This exacerbates the problem of low traffic to long-tail categories, creating a vicious cycle. This problem becomes more serious for newly built categories because of business development. Figure~\ref{fig:long-tail} illustrates the distribution of query rank versus search count. According to Zipf's law, most queries show long-tail phenomena, which makes these models hard to generalize due to a lack of training data. 

Furthermore, the posterior label of queries with similar semantics is unstable due to the randomness of user clicks. For example, when the user searches for "earphones", they may click on labels such as "Headset" and "Second-hand headset". However, if another user inputs a similar search query, such as "white earphones", the clicked labels may change to "Bluetooth earphones" or "Gaming earphones". Even though the categories of "Headset" and "Second-hand headset" also offer white earphones, they are not clicked by customers, thus not presented at the labels of the query "white earphone". This instability makes the model very sensitive to input, leading to an unstable and incomplete recall of categories. Since downstream product retrieval relies on category outcomes, an incomplete category recall cascades into relevant products not being retrieved, thereby impacting the user's purchase experience and business revenue.

To address the aforementioned challenges simultaneously, we proposed a semi-supervised multi-channel graph convolutional network. To begin, we obtain the co-occurrence relations between categories by counting the frequency of category pairs in the training samples and obtain the similarity relations between categories through the semantic relevance between categories. Despite the limited number of training samples for tail categories, tail categories are easily connected to their relevant popular categories by the co-occurrence or semantic similarity relations. These relations facilitate the transfer of gradients from samples with popular categories to tail categories, resulting in more effective representation training for long-tail categories and compensating for the drawbacks of posterior labels. Subsequently, we use a multi-channel GCN to model both relations, which enables the model to learn similar representations for the categories with higher relevance. Finally, we calculate the relevance score between the query and categories, treating it as a semi-supervised label, and then fuse it with the clicked label to calculate loss. In this way, SMGCN can use both relations between categories and the semantic similarity between query and categories as prior information to compensate for the drawbacks of the posterior data and improve the recall rate of relevant categories.

The contributions of this paper can be summarized as follows:
\begin{itemize}
\item We propose a novel and practical strategy that explicitly extends category information and utilizes the similarity score between the query and categories to augment the posterior label.

\item We design an effective model SMGCN that comprehensively leverages the relations between categories and the semantic similarity between query and categories to overcome the shortcomings of the posterior data to improve the recall rate of relevant categories.

\item The effectiveness of SMGCN has been confirmed through extensive offline experiments on two large-scale real-world datasets and online A/B test experiments. It has been deployed in production at an e-commerce platform and serves hundreds of millions of requests every day. SMGCN brings great commercial value and is a practical and robust solution for large-scale query intent classification services.
\end{itemize}

\section{Related Work}

\subsection{Multi-label Classification}
Conventional multi-label classification methods can be broadly categorized into two main types: problem transformation and algorithm adaptation methods. Problem transformation methods are multi-label techniques that transform the multi-label problem into multiple single-label problems~\cite{tsoumakas2007random,tsoumakas2009mining}, while algorithm adaptation methods~\cite{zhang2007ml,read2011classifier} focus on adapting existing algorithms to tackle the multi-label challenges.

In recent years, deep learning methods, such as RCNN~\cite{lai2015recurrent}, and XML-CNN~\cite{liu2017deep}, have been applied to capture contextual information of the document for multi-label text classification. Some seq2seq-based techniques~\cite{you2019attentionxml,xiao2019label,du2019explicit}, like MLC2Seq~\cite{nam2017maximizing} and SGM~\cite{yang2018sgm} have used an RNN to encode the text and an attention-based RNN decoder to generate predicted labels sequentially to learn the dependency of different labels. Additionally, LSAN~\cite{xiao2019label}, and LEAM~\cite{wang2018joint}, have explored label-specific attention mechanisms to capture the interactions between words and labels to learn better representations for labels and measure the compatibility of word-label pairs.

While these methodologies have demonstrated auspicious outcomes in various benchmark assessments, their applicability within industrial domains encounters distinct challenges. Industrial training datasets frequently exhibit class imbalance, and the stability of data labels remains precarious because of the randomness of user behavior. Consequently, their efficacy may be significantly undermined if they were to be employed directly in the context of online E-commerce applications.

\begin{figure*}[!htbp]
    \centering
    \includegraphics[scale=0.6]{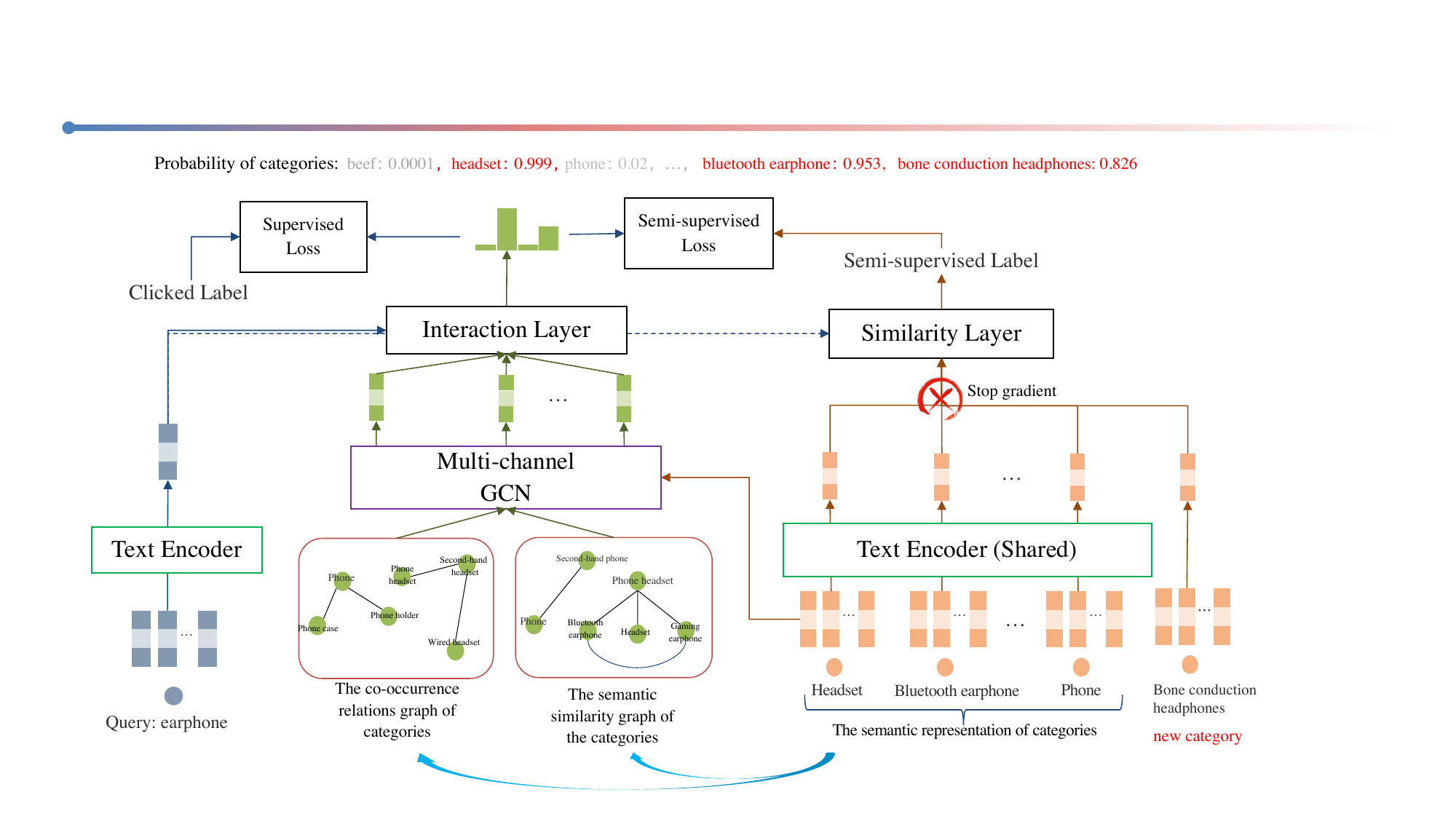}
    \caption{Semi-supervised Multi-channel Graph Convolutional Network.}
    \label{model_structure}
\end{figure*}

\subsection{Query Intent Classification}
Early query intent classification mainly focuses on mining the click graphs~\cite{li2008learning} or click-through logs~\cite{ashkan2009classifying} to improve the accuracy of prediction on the websites. In recent years, with the development of e-commerce applications and deep learning, convolutional network~\cite{hashemi2016query}, LSTM~\cite{sreelakshmi2018deep}, attention-based models~\cite{zhang2021modeling} have been exploited to learn the semantics of queries and capture fine-grained features for intent classification. For example, PHC~\cite{zhang2019improving} investigates the correlation between query intent classification and textual similarity and proposes a multi-task framework to optimize both tasks. DPHA~\cite{zhao2019dynamic} utilizes a label graph-based neural network and soft training with correlation-based label representation. MMAN~\cite{yuan2023multi} extracts features from the char and semantic level from a query-category matrix to mitigate the gap in the expression between informal queries and categories. 

While these methods considered the characteristics of the query of the industrial application, most of them rely on click logs to construct training data and suffer the problems of imbalanced categories and low recall rates of relevant categories. This paper endeavors to address these issues by leveraging two key factors: the inter-category relationships and the semantic affinity between queries and categories. By incorporating these elements as prior information, we aim to mitigate the limitations associated with the posterior data and enhance the recall performance concerning relevant categories.

\section{Model}
In this section, we first formally define the query intent classification task. Then, we describe different modules of SMGCN in detail and analyze the influence of the model during the training and predicting process.

\subsection{Problem Statement}
Suppose the query inputted by users on the E-commerce applications, has $q = [x_1, x_2, \ldots, x_{L_q}]$ characters. Each category $n_i$ has a category name and a series of product words, and $|C|$ denotes the total amount of leaf categories. The products belong to one of the leaf categories. The query classification task requires models to assign a subset $y$ of categories from all leaf categories to $q$. Our target is to learn a classification model $f(\cdot,\cdot)$. For any input query $q$, the model $f(q, [n_1, .., n{_C}])$ can select relevant categories from the label set. For a clear definition, throughout the rest of this paper, bold lowercase letters represent vectors.

\subsection{Overview}
Figure~\ref{model_structure} illustrates the components of the SMGCN model, which is mainly composed of three modules: (1) query and category representation learning module, (2) semi-supervised label generation module, and (3) multi-channel graph learning module. Specifically, the query and category representation learning module describes the mapping of query or category sequence from word embedding into the same semantic space; the semi-supervised label generation module illustrates why the model needs semi-supervised labels and how to utilize the pseudo-label to facilitate model training; the multi-channel graph learning module defines two kinds of relations of categories and introduces how to fuse both kinds of relations to learn better category embeddings. Finally, query and category embeddings are fed to a classifier to predict the user's intent.

\subsection{Learning query and category representation}
Query and categories are the basic input of the model. To learn good semantic representations of them, we project the query and category onto the same vector space. BERT~\cite{kenton2019bert,liu2019roberta} has gained widespread industrial applications, so we employ BERT as the encoder for both the query and categories. 

To learn the semantics of the product categories, the category character sequence is comprised of two distinct components: (1) the category name $n = [n_1, n_2, \ldots, n_{C}]$ where $C$ denotes the number of categories; (2) the selected core product words $m = [m_1, m_2, \ldots, m_{L_m}]$ for $n_i$, where ${L_m}$ denotes the number of product words.

Once we have obtained the high-quality product words, we concatenate them with category names and subsequently feed them into BERT to encode category representation. To project queries and categories onto a common semantic space, the query and category share the same BERT model, which can be expressed as follows:
\begin{equation}
\begin{split}
& \mathbf{Q}_i = \mathrm{BERT_{CLS}}([x_1, x_2, \ldots, x_{L_q}])   \,, \\
& \mathbf{C}_j = \mathrm{BERT_{CLS}}([n_j, m_1, \ldots, m_{L_m}]) \,, \\
\end{split}
\end{equation}
where $ \mathrm{BERT_{CLS}}$ is the ``CLS'' representation of the last layer of BERT; $\mathbf{Q}_i \in \mathbb{R}^{1 \times d}$ and $\mathbf{C} \in \mathbb{R}^{|C| \times d} $  denote the token embedding matrix of query and category, respectively.

\subsection{Semi-supervised label generation}
Most existing methods rely on user click behavior to generate training samples, but long-tail categories struggle to obtain traffic and user clicks compared to popular categories. Additionally, user click behavior tends to be random and unstable for queries with similar semantics due to individual preferences and varying demands. As a result, the posterior labels are highly imbalanced and unstable, leading to inadequate performance for long-tail categories and incomplete category recall.

To compensate for the drawbacks of the posterior label,  we calculate the similarity score between the query and categories to treat it as a semi-supervised label. Then, we fuse it with the label clicked by the user to calculate loss as the final label. Specifically, 
\begin{equation}
\begin{split}
& \mathbf{s}_i = \mathop{stop\_grad ( \frac{\mathbf{Q}_i \mathbf{C}^T}{\lVert Q_i\rVert   \lVert C\rVert} )  }  \,,  \\
& \mathbf{y}^{semi}_{ij} = \begin{cases}
\mathbf{s}_{ij} & \text{if } \mathbf{s}_{ij} \geq \tau \\
0  & \text{if } \mathbf{s}_{ij} < \tau   \\
\end{cases}         \,,  \\
\end{split}
\end{equation}
where $\mathbf{s}_i \in \mathbb{R}^{1 \times |C|}$, is the relevance scores between query $q_i$ and all categories. $\tau$ is the threshold to filter the categories with low scores. $\mathbf{y}^{semi}_{ij}$ is the semi-supervised label. For example, referring to Figure~\ref{model_structure}, although the new category "Bone conduction headphones" did not have click records below query "earphone", they are semantically highly related and should be recalled. This connection can be expressed by the $\mathbf{y}^{semi}_{ij}$  and influences model training.

Both query and label encoders use the same text encoder, but their word distribution is different. If the gradient of the semi-supervised signal is fed to the semi-supervised label generation module, a circular dependency may arise, which could ultimately result in the model collapse. To avoid this issue, we disable the gradient feedback of this branch and solely rely on the gradient of semi-supervised labels to guide the training of the query intent classification module.

\subsection{Multi-channel graph learning}
Subsequently, we will introduce how the model leverages the relations among categories as prior information to compensate for the drawbacks of the posterior labels. 

\subsubsection{Graph construction}
Firstly, we obtain the co-occurrence relations between categories by counting the co-occurrence times of categories in the training samples. Then, we compute the conditional probability of two categories and obtain the adjacency matrix $\mathbf{A}^{coo}$: 
\begin{equation}
\begin{split}
& \mathbf{A}^{coo}_{ij} =\frac{N(c_i, c_j)}{N(c_i)}  \,,  \\
\end{split}
\end{equation}
where $N(c_i, c_j)$ is co-occurrence times of category $c_i$ and $c_j$ and $N(c_i)$ denotes the number of occurrences of category $c_i$.

Additionally, we obtain the semantic similarity relations between categories by computing the cosine similarity of every pair of categories:
\begin{equation}
\begin{split}
& \mathbf{a}_{ij} = \frac{\mathbf{C}_i \mathbf{C}^T_j}{\lVert C_i\rVert   \lVert C_j\rVert} \,,  \\
& \mathbf{A}^{sim}_{ij} = \begin{cases}
\mathbf{A}^{sim}_{ij} & \text{if } \mathbf{a}_{ij} \geq \alpha \\
0  & \text{if } \mathbf{a}_{ij} < \alpha   \\
\end{cases}         \,,  \\
\end{split}
\end{equation}
where $\alpha$ is the threshold to filter the edges with low relevance scores.

The two correlation matrices obtained from different scales cannot be merged directly, so it is necessary to normalize $\mathbf{A}^{coo}$ and $\mathbf{A}^{sim}$ respectively. The normalization method~\cite{kipf2017semi} is formalized as follows:
\begin{align}
\widehat {\mathbf{A}} &= \mathbf{D}^{-\frac{1}{2}}\mathbf{AD}^{-\frac{1}{2}} \,,
\end{align}
where $\mathbf{D}$ is a diagonal degree matrix with entries $D_{ij} = \Sigma _{j}\mathbf{A}_{ij}$.

Next, we merge the two correlation matrices after normalization: 
\begin{align}
\mathbf{A} = [\widehat {\mathbf{A}}^{coo}; \widehat {\mathbf{A}}^{sim}] \,,
\end{align}
where $\mathbf{A} \in \mathbb{R}^{ 2 \times |C| \times |C|}$ is the fused adjacency matrix.

\subsubsection{Graph learning}
GCN is applied to generate nodes' representation by aggregating neighborhood information. The layer-wise propagation rule of a multi-layer GCN is as follows:
\begin{equation}
\mathbf{H}^{l+1} = \mathbf{LeakyReLU}\left( \mathbf{A}\mathbf{H}^{l}\mathbf{W}^{l}\right) \,,
\end{equation}
where $\mathbf{H}^{l} \in \mathbb{R}^{|C| \times d}$ is in the $l^{th}$ layer (where $|C|$ denotes the number of nodes, $d$ is the dimensionality of node features) and $\mathbf{H}^{l+1}$ is the enhanced node features. $\mathbf{W}^l \in \mathbb{R}^{d \times d'}$ is a transformation matrix to be learned.

Despite the limited number of training samples for tail categories, tail categories are easily connected to their relevant hot categories by the co-occurrence or semantic similarity relations. These relations facilitate the transfer of gradients from samples with hot categories to tail categories, resulting in more effective representation training for long-tail categories and compensating for the drawbacks of posterior labels.

\begin{table*}[!htbp]
    \centering
    \caption{Dataset statistics.}
    \setlength{\tabcolsep}{6mm}{
        \begin{tabular}{c|ccc|ccc}
            \toprule
            \multirow{2}{*}{\textbf{Statistics}} & \multicolumn{3}{c}{\textbf{Intent Data}} & \multicolumn{3}{c}{\textbf{Category Data}}  \\ 
            &\textbf{Train} &\textbf{Validation} &\textbf{Test}  &\textbf{Train} &\textbf{Validation} &\textbf{Test}  \\   
            \hline
            \hline 
            Queries  & 67,450,702 & 20,0000  & 31,792 & 113,686,150 & 20,0000 & 33,960 \\
            \hline
            Avg. chars & 7.63 & 5.00   &  8.36    & 8.50 & 6.53   &  6.02 \\
            \hline
            Total Labels  & 1,605 & 1,605    & 1,605    & 6,634 & 6,634   &  6,634 \\
            \hline
            Avg. \# of labels  & 1.04  & 1.67 &  1.91    & 1.52 & 2.05   &  5.33  \\
            \hline
            Min. \# of labels & 1 & 1   &  1  & 1 & 1   &  1  \\
            \hline
            Max. \# of labels & 7 & 3  &  16  & 16 & 13  &  20 \\
            \bottomrule
        \end{tabular}
    }
    \label{tab:datset}
\end{table*}

\subsection{Training and inference}
Finally, we obtain query representation $\mathbf{q}_i \in \mathbb{R}^{1 \times d}$ and the final representations of categories $\mathbf{H} \in \mathbb{R}^{|C| \times d}$. Specifically, we introduce the nonlinear transformation layer which is defined as:
\begin{equation}
    \widehat{\mathbf{y}}_i = \mathbf{sigmoid}(\mathbf{q}_i \mathbf{H}^T + \mathbf{b}) \,,
\end{equation}
where $\mathbf{b} \in \mathbb{R}^{1 \times |C|}$ is the bias, and $\widehat{\mathbf{y}}_i  \in \mathbb{R}^{1 \times |C|}$ is the predicted labels of query $q_i$.

To optimize the model and use the posterior and semi-supervised labels, we fuse them as follows:
\begin{equation}
\begin{split}
& \mathbf{y} = \mathbf{y}^{click} + \mathbf{y}^{semi}  \,,   \\
& \mathbf{y}_{i} = \begin{cases}
\mathbf{y}_{i} & \text{if } \mathbf{y}_{i} \le 1.0 \\
1.0  & \text{else}   \\
\end{cases}         \,,  \\
\end{split}
\end{equation}
where $\mathbf{y}_i^{click}$ is the one-hot encoding of clicked labels of query $q_i$, and the value range of $\mathbf{y}_{i} $ is $\mathbf{y}_{i} \in [0, 1]$.

In this paper, we use the binary cross-entropy loss as the objective to train the model :
\begin{align}
\mathcal{L} &= - \sum ^{N}_{j=1} \sum ^{|C|}_{i=1} \mathbf{y}_i \log \left( \widehat{\mathbf{y}}_i \right) + \left( 1 - \mathbf{y}_i \right) \log \left( 1 - \widehat{\mathbf{y}}_i \right) \,,
\end{align}
where $N$ is number of samples, $\mathcal{L}$ is the final loss function.

\section{Experiment}
In this section, we will discuss the offline and online experiments in detail. We first introduce the datasets and the evaluation metrics used in this paper. Then, we analyze the experiment results by several fair comparisons with strong baselines. After that, we deeply investigate the effect of different modules of the SMGCN model. Subsequently, we present the online performance of the model on the JD search engine and further analyze the influence of different modules. Finally, we explore the influence of hyper-parameters.

\subsection{Dataset}
\label{sec:Dataset}
To evaluate the effectiveness and generality of the proposed model, we conducted a series of experiments on two large-scale real-world datasets collected from users' click logs on the JD application. The statistics of the datasets are listed in Table~\ref{tab:datset}. Specifically, 
\begin{itemize}
    \item \textbf{Category Data}: We randomly sample queries and corresponding clicked products from search logs over one month. The clicked products' category are treated as the query's intent. The clicked frequency of the product is treated as the frequency of the category. To filter unreliable categories, we normalize the frequency of the category and compute the cumulative distribution function (CDF) of the category's probability. When $CDF > 0.95$, the rest of the categories with low probabilities are removed. 
    
    \item \textbf{Intent Data}: The e-commerce platform defines a hierarchical intent architecture that contains more than 1000 intents of users by many experts. The categories of the query are mapped into intent domains, which form the Intent data. The training data is also mined with identical rules as the category data from user historical click logs. Different from it, the test dataset is annotated by the experts in each domain. 
\end{itemize}

\subsection{Baseline Models}
We compare SMGCN with several strong baseline models, including widely-used multi-label classification methods, such as XML-CNN, and LSAN, and query intent classification models, such as PHC, DPHA, and MMAN. The detailed introductions are listed as follows:

(1) Multi-label text classification baselines: 
\begin{itemize}
    \item \textbf{RCNN}~\cite{lai2015recurrent}: It captures contextual information with the recurrent and convolutional structure for text classification. 
    \item \textbf{XML-CNN}~\cite{liu2017deep}: It is a CNN-based model, which combines the strengths of CNN models and goes beyond the multi-label co-occurrence patterns. 
    \item \textbf{LEAM}~\cite{wang2018joint}: It is a label-embedding attentive model, which embeds the words and labels in the same space, and measures the compatibility of word-label pairs. 
    \item \textbf{LSAN}~\cite{xiao2019label}: It is a label-specific attention network that uses document and label text to learn the label-specific document representation with the self- and label-attention mechanisms.
\end{itemize}

(2) Query intent classification baselines: 
\begin{itemize}
    \item \textbf{CNN}~\cite{hashemi2016query}: It proposes a convolutional neural network (CNN) to extract query vector representations as the features for the query classification.
    \item \textbf{PHC}~\cite{zhang2019improving}: It investigates the correlation between query intent classification and textual similarity and proposes a multi-task framework to optimize both tasks. 
    \item \textbf{BERT}~\cite{kenton2019bert}: We use the pre-trained BERT-base~\footnote{https://tfhub.dev/tensorflow/bert\_zh\_L-12\_H-768\_A-12/4} delivered by google, and fine-tune it on the training set to predict the user's intent. 
    \item \textbf{DPHA}~\cite{zhao2019dynamic}: It contains a label graph-based neural network and soft training with correlation-based label representation. 
    \item \textbf{MMAN}~\cite{yuan2023multi}: It is a BERT-based model that extracts features from the char and semantic level from a query-category interaction matrix to mitigate the gap in the expression between informal queries and categories.
\end{itemize}

\begin{table*}[!htbp]
  \caption{
        The experimental results compared to multi-label text classification and query intent classification models. 
  }
  \centering
  \label{tab:experiment}
  \setlength{\tabcolsep}{1.8mm}{
      \begin{tabular}{c|ccc|ccc|ccc|ccc}
                \toprule
                
                \multirow{3}{*}{\textbf{Models}}  & 
                \multicolumn{6}{c|}{\textbf{Intent Data}} & \multicolumn{6}{c}{\textbf{Category Data}} \\
                
                & \multicolumn{3}{c|}{\textbf{Micro}}  
                & \multicolumn{3}{c|}{\textbf{Macro}} 
                & \multicolumn{3}{c|}{\textbf{Micro}}  
                & \multicolumn{3}{c}{\textbf{Macro}}  \\
                
                &\textbf{Precision} &\textbf{Recall} &\textbf{F1}
                &\textbf{Precision} &\textbf{Recall} &\textbf{F1}
                &\textbf{Precision} &\textbf{Recall} &\textbf{F1}
                &\textbf{Precision} &\textbf{Recall} &\textbf{F1} \\
                \midrule
                \midrule
                RCNN      & 79.20 & 34.72 & 48.28 & 53.72 & 23.39 & 30.37  & 84.32 & 27.26 & 41.20  & 39.86 & 16.49 & 21.02 \\
                XML-CNN   & 78.66 & 32.09 & 45.58 & 50.33 & 20.76 & 27.24  & 86.95 & 24.60 & 38.34  & 40.50 & 15.44 & 20.16  \\
                LEAM      & 76.22 & 37.21 & 50.01 & 55.11 & 25.72 & 32.40  & 76.79 & 26.68 & 39.60  & 39.40 & 17.19 & 21.31  \\
                LSAN      & 76.46 & 34.96 & 47.98 & 54.47 & 25.12 & 31.71  & 86.39 & 23.66 & 37.15  & 44.69 & 17.79 & 22.84  \\
                
                \midrule 
                CNN       & 77.36 & 37.85 & 50.83  & 55.71 & 26.10 & 32.89  & 88.18 & 24.11 & 37.86  & 39.27 & 14.36 & 18.94   \\
                PHC       & 77.94 & 36.03 & 49.28 & 56.12 & 25.43 & 32.33  & 82.84 & 27.33 & 41.10  & 42.20 & 18.39 & 22.97  \\
                DPHA      & 77.22 & 36.91 & 49.94 & 55.09 & 25.74 & 32.53  & 87.29 & 22.49 & 35.76  & 36.08 & 13.11 & 17.26  \\
                MMAN      & 79.26 & 38.96 & 52.24 & 56.27 & 26.32 & 33.36  & 82.05  &  32.57  &  46.63  & 57.41  &  28.26   & 34.68  \\
                
                \midrule
                \textbf{SMGCN}  & 75.83 & 49.91  & 59.72  & 63.18  & 43.90  & 48.54   & 82.51 & 40.05 & 53.92  & 55.83 & 35.62 & 40.15 \\
                \bottomrule
        \end{tabular}
    }
\end{table*}

\subsection{Evaluation Metrics}
Query intent classification is essentially a multi-label text classification task. Thus, following the settings of previous work~\cite{zhang2021modeling,yuan2023multi}, we report the micro and macro precision, recall, and F1-score of the models as the metrics to evaluate their performance. The definitions of these metrics are listed as follows:
\begin{itemize}
    \item {Micro-Precision / Recall / F1}: The calculation of the micro average metric requires aggregating the contributions of all labels to compute the average micro score. The categories with more samples have an advantage over other categories.
    
    \item {Macro-Precision / Recall / F1}: The macro average metric computes the score independently for each label and then takes the average as the final score. Thus, each category has a similar contribution to the overall score.
\end{itemize}

\subsection{Experiment Settings}
We implement the models based on the Pytorch framework. The dimensionality of the embedding of BERT is 768. We use a 2-layer GCN to learn the category embeddings of two graphs, and the dimensionality of embedding is 768. We use Adam algorithm~\cite{kingma2014adam} with a learning rate of $1e^{-4}$. The max length of the query is set to 16. The threshold of labels is set to 0.5. The threshold $\tau$ is set to 0.8 and $\alpha$ is set to 0.65 according to the result of the grid search. The model training should use a warm start strategy and the threshold $\tau$ is gradually decreased to 0.8 as the training.

To overcome the overfitting, we use the dropout strategy with a dropout rate of 0.5. The maximum training epoch is set to 20, and the batch size of the training set is set to 1024. We select the best parameter configuration based on the performance of the validation set and evaluate the configuration on the test set.

\subsection{Offline Evaluation}

\subsubsection{Offline performance}
The experimental results are shown in Table~\ref{tab:experiment}. Overall, the experimental results indicate that SMGCN significantly outperforms all baselines on two large-scale real-world datasets. Specifically, we have the following observations:

(1) For the multi-label text classification baselines (i.e., RCNN, XML-CNN, LEAM, and LSAN), it is obvious that SMGCN outperforms them by a significant margin on two large-scale datasets. These methods mainly focus on learning better query and label representations but ignore the complexity of real industrial applications. Industrial training datasets frequently exhibit class imbalance, data distribution is often dominated by popular categories and the stability of data labels remains precarious because of the randomness of user behavior. Consequently, their efficacy may be significantly undermined if they were to be employed directly in the context of online E-commerce applications.

(2) Compared with recently proposed query intent classification methods (i.e., CNN, PHC, DPHA, and MMAN), SMGCN also achieves better performance on both datasets. As the results are shown in the table, the recall of relevant categories obtains nearly 10\% improvement whether from the micro or macro view on both datasets. The improvement achieved by macro metrics is larger than that of micro metrics, which further proves that the SMGCN model has a greater improvement and better effect on long-tail categories. While baseline models consider the matching features between query and label, none of them treat this interaction as a supervised signal to train the model. Only adding interaction features is not enough to change the distribution of imbalanced categories. This paper addresses this problem by utilizing inter-category relations and treating the semantic affinity between queries and categories as supervised soft labels. By incorporating these elements as prior information, we mitigate the problems brought by the posterior data and enhance the recall performance concerning relevant categories.

In conclusion, SMGCN achieves significant improvement over all baselines in terms of micro-F1 and macro-F1 scores. This confirms that using the relations among categories, as well as the semantic similarity between queries and categories, is beneficial for overcoming the limitations associated with posterior data and for improving the recall rate of the categories.

\begin{table*}[!htbp]
  \caption{Ablation study of the proposed model SMGCN.}
  \centering
  \label{ablation_study}
  \setlength{\tabcolsep}{1.5mm}{
      \begin{tabular}{c|ccc|ccc|ccc|ccc}
            \toprule
            \multirow{3}{*}{\textbf{Models}}  & 
            \multicolumn{6}{c|}{\textbf{Intent Data}} & \multicolumn{6}{c}{\textbf{Category Data}} \\
            
            & \multicolumn{3}{c|}{\textbf{Micro}}  
            & \multicolumn{3}{c|}{\textbf{Macro}} 
            & \multicolumn{3}{c|}{\textbf{Micro}}  
            & \multicolumn{3}{c}{\textbf{Macro}}  \\
            
            &\textbf{Precision} &\textbf{Recall} &\textbf{F1}
            &\textbf{Precision} &\textbf{Recall} &\textbf{F1}
            &\textbf{Precision} &\textbf{Recall} &\textbf{F1}
            &\textbf{Precision} &\textbf{Recall} &\textbf{F1} \\

            \midrule
            \textbf{SMGCN}  & 75.83 & 49.91  & 59.72  & 63.18  & 43.90  & 48.54   & 80.51 & 40.05 & 53.49  & 55.83 & 35.62 & 40.15 \\
            w/o. simi. graph       & 79.54 & 43.25  & 56.03  & 64.30  & 35.96  & 42.98   & 81.28 & 37.54 & 51.36  & 57.24 & 32.17 & 37.86 \\
            w/o. coo. graph       & 76.83 & 45.69  & 57.30  & 49.24  & 31.03  & 34.83   & 80.12 & 38.58 & 52.08  & 56.18 & 34.05 & 39.52  \\
            w/o. graph       & 80.12 & 41.18  & 54.40  & 54.64  & 37.35  & 41.26  & 83.05  &  35.17  &  49.42  & 56.79  &  30.90   & 36.37 \\
            BERT            & 81.28 & 37.59  & 51.41  & 51.63  & 29.97  & 36.84  & 82.83  &  31.99  &  46.15  & 56.72  &  27.80   & 33.80  \\
            \bottomrule
        \end{tabular}
    }
\end{table*}

\subsubsection{Ablation study}
To further figure out the relative importance of each module in the proposed model, we perform a series of ablation studies over the different components of SMGCN. Three variants of SMGCN are listed below: 

\begin{itemize}
    \item \textbf{w/o simi. graph}: Removing the graph constructed through the semantic similar relations between category pairs. Only use the co-occurrence graph and semi-supervised strategy for query intent prediction.
    
    \item \textbf{w/o coo. graph}: Removing the graph constructed through the co-occurrence relations between category pairs and using the similarity graph with the semi-supervised strategy for query intent prediction.
    
    \item \textbf{w/o graph}: Removing both co-occurrence and similarity graphs only uses the semi-supervised strategy with BERT for intent prediction. 
    
    \item \textbf{BERT}: Removing all modules and only remaining BERT as text encoder for query intent classification.
\end{itemize}

The experiment results are shown in Table~\ref{ablation_study}. We can observe that:

(1) When removing the similarity graph, the performance consistently has a little drop compared with SMGCN on both datasets. A similar phenomenon can be seen when removing the co-occurrence graph, indicating that the similarity or co-occurrence graph does contain extra information that is neglected in the posterior data. 

(2) When we eliminate both similarity and co-occurrence graphs, the performance degrades more than 5\% compared with the complete SMGCN. The results indicate that both graphs play different roles in category representation learning. 

(3) After removing these three modules, we can see that the micro and macro F1 decay about 8\% compared with the complete SMGCN. This result further demonstrates that all of these components in SMGCN provide complementary information to each other, and are requisite for query intent classification.

\subsection{Online Evaluation}

\subsubsection{Online Deployment}
To reduce the response latency of online deployment, the text encoder of the SMGCN is distilled from the 12-layer BERT base to the 4-layer BERT. Moreover, it is not necessary to deploy the multi-channel GCN online. We only export the category embeddings $\mathbf{H} \in \mathbb{R}^{|C| \times d}$ produced by the multi-channel GCN. When we obtain query embedding from the text encoder, we compute the dot product between query embedding and $\mathbf{H}$ for classification. In this way, we can deploy SMGCN without adding any additional computation and latency compared with pure BERT.

\begin{figure}[t]
\centering
\includegraphics[width=9cm,height=9cm]{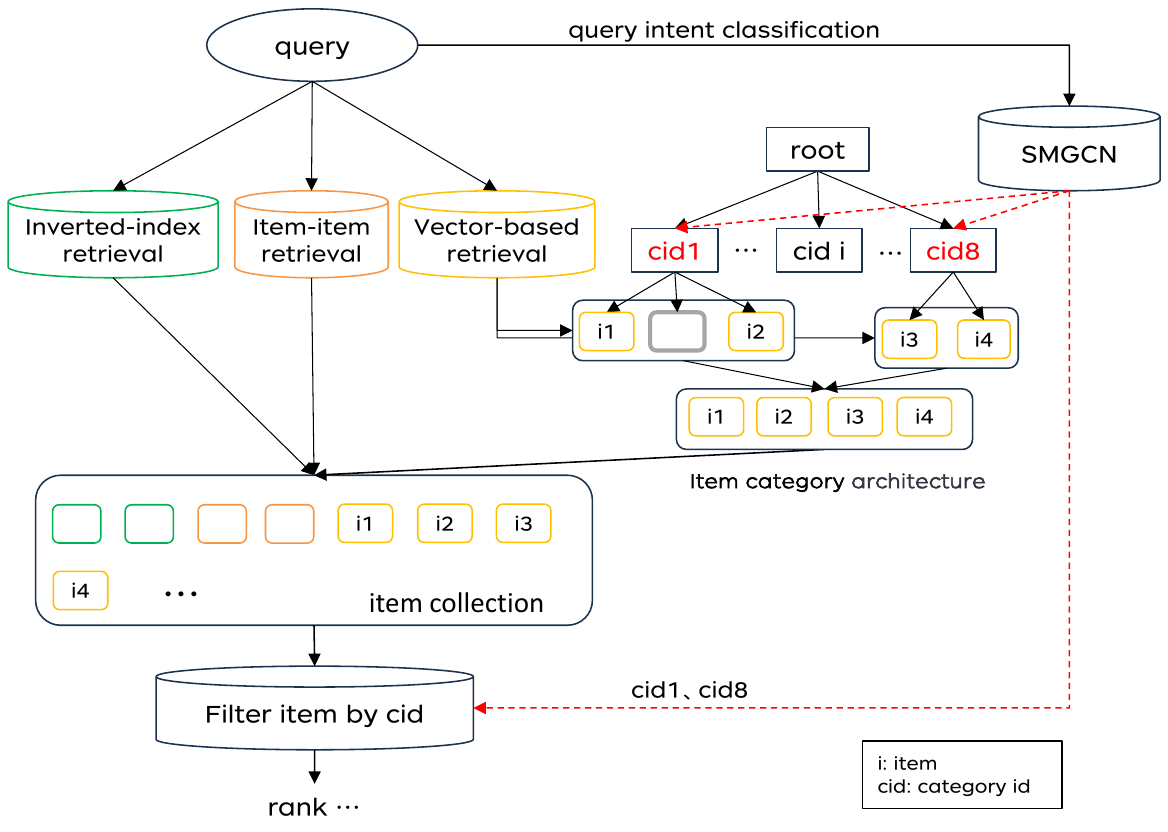}
\caption{The deployment of SMGCN and the role of category plays in the E-commerce system.}
\label{fig:system}
\end{figure}

Figure~\ref{fig:system} illustrates the role the SMGCN model plays in the JD search system. The items are organized and retrieved in an item category architecture with three levels. When a user inputs a query, the SMGCN model first classifies the user's intent and sends the categories that the user demands to downstream modules. Then, the vector-based retrieval module will retrieve relevant items below the category IDs that the SMGCN gives. The retrieved items will be integrated with items from other retrieval branches and then subjected to filtering by a sub-module designed to exclude irrelevant items that do not correspond to the user's desired categories. The filtered items will be sent to the rank module. Overall, the predicted categories by SMGCN mainly influence the vector-based retrieval branch during the product retrieval process and influence the query-item relevance computing process.

\subsubsection{Online Performance}
Before being launched in production, we routinely deployed the SMGCN online on the JD search engine and made it randomly serve 10\% traffic as the test group. For a fair comparison, we also build a base group that using the previous model (4-layer BERT) serves 10\% traffic. During the A/B testing period, we monitor the performance of SMGCN and compare it with the online model. This period lasts for at least one week. 

For online evaluation, we use some business metrics: UV value, UCVR (conversion rate of users), UCTR (click rate of users), and Diversity (category diversity of the exposed products). The specific computation of these metrics is defined as follows: 
\begin{equation}
\begin{split}
    & UV value = \frac{GMV}{UV}    \,,  \\
    & UCVR = \frac{\#Orderlines}{UV}    \,,  \\
    & UCTR = \frac{\#Clicks}{UV}    \,,  \\
\end{split}
\end{equation}
where $UV$ denotes the number of unique visitors, \#Orderlines denotes the total number of purchases made by all users on the e-commerce platform, and GMV denotes the gross merchandise value.

\begin{table}[!htbp]
  \centering
  \caption{Online improvements of the SMGCN. Improvements are statistically significant with $p < 0.05$  on paired t-test. All performances of SMGCN and its variants are compared with the online model.}
  \label{online_performance_uv_ucvr}
  \setlength{\tabcolsep}{1.5mm}{
      \begin{tabular}{c|cccc}
            \toprule
            Models &\textbf{UV value} &\textbf{UCVR}  & \textbf{UCTR}  & \textbf{Diversity}
            \\
            \midrule
            Online         & - & -  & - & -   \\
            SMGCN          & +0.79\%   & +0.65\%  & +0.20\%    & +2.87\%   \\
            w/o. simi. graph  & +0.41\%   & +0.44\%  & +0.03\%    & +2.15\%   \\
            w/o. coo. graph  & +0.38\%   & +0.41\%  & +0.08\%    & +2.18\%   \\
            w/o. graph      & +0.27\%   & +0.22\%  & +0.11\%    & +1.41\%   \\
            \bottomrule
        \end{tabular}
    }
\end{table}

The online A/B experimental results are shown in Table~\ref{online_performance_uv_ucvr}. We can observe that the category diversity of the exposed products gets a dramatic improvement compared with the base group, which means (1) the incremental categories retrieved by the SMGCN are indeed the categories the users required; (2) by increasing the recall rate of relevant categories, more related products are retrieved, making users click and buy more products, leading to more UCVR and UV value improvement; (3) as the sub-modules of the model are removed, the online performance continues to decline, which further confirms the effectiveness of the different modules of SMGCN. 

In conclusion, both the results of the offline evaluation and online A/B experiments consistently demonstrate the effectiveness, efficiency, and universality of the proposed SMGCN model.

\subsection{Parameter Sensitivity}
\begin{figure}[!htbp]
    \centering
    \subfigure[The max length of query $L_q$]{
        \label{fig:subfig:a1} 
        \includegraphics[width=4cm,height=4.1cm]{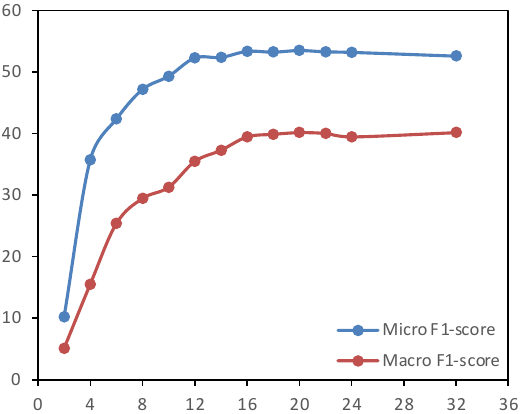}
    }
    \subfigure[The max length of category information  $L_c$]{
        \label{fig:subfig:a2} 
        \includegraphics[width=4cm,height=4.1cm]{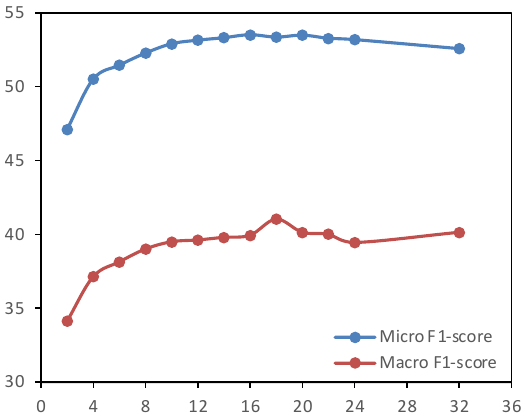}
    }
    
    \subfigure[The threshold $\tau$]{
        \label{fig:subfig:b1} 
        \includegraphics[width=4cm,height=4cm]{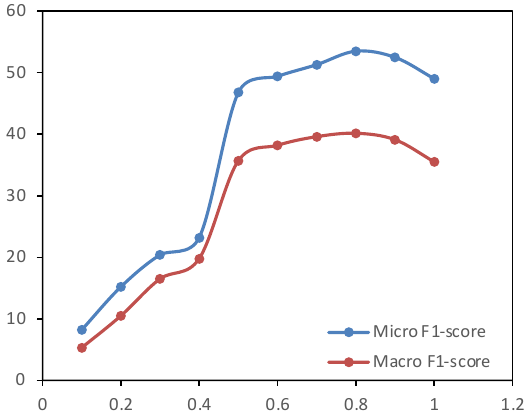} 
    }
    \subfigure[The threshold $\alpha$]{
        \label{fig:subfig:b2} 
        \includegraphics[width=4cm,height=4cm]{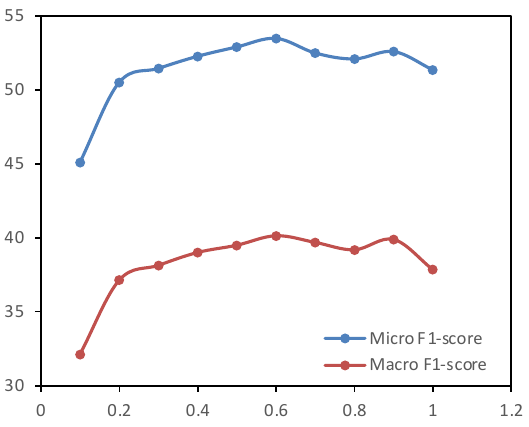} 
    }
    \caption{The performance of the proposed framework on the majority and minority styles. }
    \label{parameter_sensitivity_analysis}
\end{figure}

Four major hyper-parameters may influence the performance: (1) The maximum length of query and category; (2) the threshold $\tau$ and $\alpha$. We conduct some sensitivity analysis experiments to study how different choices of hyper-parameters influence the performance of the SMGCN. The results are shown in Figure~\ref{parameter_sensitivity_analysis}. Due to space limitations, we only show the results on the category dataset.

\begin{itemize}
    \item Impact of the maximum length. Figure~\ref{parameter_sensitivity_analysis} (a) and (b) illustrate the performance with different query and category lengths. The length has a significant influence on the prediction performance. When the tweet is too short, it cannot provide enough information for classification. Therefore, the performance improves as the growth of length. We observed that the best max length of the query is about 16 and the best max length of the category input is about 20. 

    \item Impact of threshold. Figure~\ref{parameter_sensitivity_analysis} (c) and (d) illustrate the performance of SMGCN with different $\tau$ and $\alpha$. $\tau$ determines how many soft labels would add to loss and $\alpha$ decides how many semantic similar edges of categories would be remained. As shown, a low threshold $\tau$ or $\alpha$ significantly influences the performance of the model because it brings too much noise. Moreover, a high threshold will filter too many useful connections of categories and also influence the performance of the model. We can observe that SMGCN achieves the best performance when $\tau = 0.8$ and $\alpha = 0.6$. 
\end{itemize}

\section{Conclusion and Future Work}
This paper proposes a semi-supervised multi-channel graph convolutional network to address the challenges of category imbalance and incomplete recall of categories. SMGCN extends category information and enhances the posterior label by utilizing the similarity score between the query and categories. Additionally, it leverages the co-occurrence and semantic similarity relations among categories to strengthen the relations between labels and weaken the influence of posterior label instability. Offline and online A/B experiments demonstrate significant improvements over the state-of-the-art methods. Moreover, the proposed approach has been deployed in real-world applications and has brought great commercial value, confirming its practicality and robustness for large-scale query intent classification services.

In future work, we aim to investigate the use of external knowledge, such as the taxonomic hierarchy of categories and product information, to comprehensively model category representations and further enhance the model's performance.

\clearpage

\begin{spacing}{1.2}
\bibliographystyle{ACM-Reference-Format}
\balance
\bibliography{sample-base}
\end{spacing}




\end{document}